\documentclass[sigconf, nonacm, pdfa]{acmart}
\usepackage[a-2b]{pdfx}

\usepackage[capitalize,nameinlink]{cleveref}
\usepackage{enumitem}
\usepackage{graphicx}
\usepackage{listings}
\usepackage{subcaption}
\usepackage{tabularx}
\usepackage{tikz}
\usepackage{xcolor}
\usepackage{xspace}
\usepackage{verbatim}

\newcommand*\numcircledmod[1]{\raisebox{.5pt}{\textcircled{\raisebox{-.9pt} {#1}}}}

\newcommand{\sys}{\textsc{Credo}\xspace}
\newcommand{\sysbf}{\textbf{\textsc{Credo}}\xspace}

\newcommand\vldbdoi{10.14778/3827998.3828054}
\newcommand\vldbpages{4518 - 4521}
\newcommand\vldbvolume{19}
\newcommand\vldbissue{12}
\newcommand\vldbyear{2026}
\newcommand\vldbauthors{\authors}
\newcommand\vldbtitle{\shorttitle}
\newcommand\vldbavailabilityurl{https://github.com/brown-db/credo}
\newcommand\vldbpagestyle{empty}

\begin{document}
\title[Credo: Declarative Control of LLM Pipelines via Beliefs and Policies]{Credo: Declarative Control of LLM Pipelines \\ via Beliefs and Policies}

\author{Duo Lu}
\affiliation{\institution{Brown University}}
\email{duo_lu@brown.edu}

\author{Andrew Crotty}
\affiliation{\institution{Northwestern University}}
\email{andrew.crotty@northwestern.edu}

\author{U\u{g}ur \c{C}etintemel}
\affiliation{\institution{Brown University}}
\email{ugur_cetintemel@brown.edu}

\begin{abstract}
AI assistants are becoming commonplace in domains that require stateful decision-making in continuously evolving environments.
As such, correctness depends on an assistant's ability to flexibly adapt by incorporating new evidence or revising prior conclusions.
However, existing approaches rely on either static control flow (e.g., fixed pipelines) or unreliable agentic reasoning.

We therefore introduce \sys, a database-backed semantic control plane for LLM pipelines that represents state as high-level \textit{beliefs} and guides execution through declarative \textit{policies} defined in terms of those beliefs.
Through this design, \sys enables adaptivity based on runtime signals while also simplifying auditing and debugging.
This demo will illustrate these concepts using an LLM pipeline built for a financial question answering benchmark.
\end{abstract}

\maketitle

\pagestyle{\vldbpagestyle}
\begingroup\small\noindent\raggedright\textbf{PVLDB Reference Format:}\\
\vldbauthors. \vldbtitle. PVLDB, \vldbvolume(\vldbissue): \vldbpages, \vldbyear.\\
\href{https://doi.org/\vldbdoi}{doi:\vldbdoi}
\endgroup
\begingroup
\renewcommand\thefootnote{}\footnote{\noindent
This work is licensed under the Creative Commons BY-NC-ND 4.0 International License. Visit \url{https://creativecommons.org/licenses/by-nc-nd/4.0/} to view a copy of this license. For any use beyond those covered by this license, obtain permission by emailing \href{mailto:info@vldb.org}{info@vldb.org}. Copyright is held by the owner/author(s). Publication rights licensed to the VLDB Endowment. \\
\raggedright Proceedings of the VLDB Endowment, Vol. \vldbvolume, No. \vldbissue\ %
ISSN 2150-8097. \\
\href{https://doi.org/\vldbdoi}{doi:\vldbdoi} \\
}\addtocounter{footnote}{-1}\endgroup

\ifdefempty{\vldbavailabilityurl}{}{
\vspace{.3cm}
\begingroup\small\noindent\raggedright\textbf{PVLDB Artifact Availability:}\\
The source code, data, and/or other artifacts have been made available at \url{\vldbavailabilityurl}.
\endgroup
}

\section{Introduction}
\label{sec:intro}

\begin{figure}[t]
\centering
\includegraphics[width=0.96\linewidth]{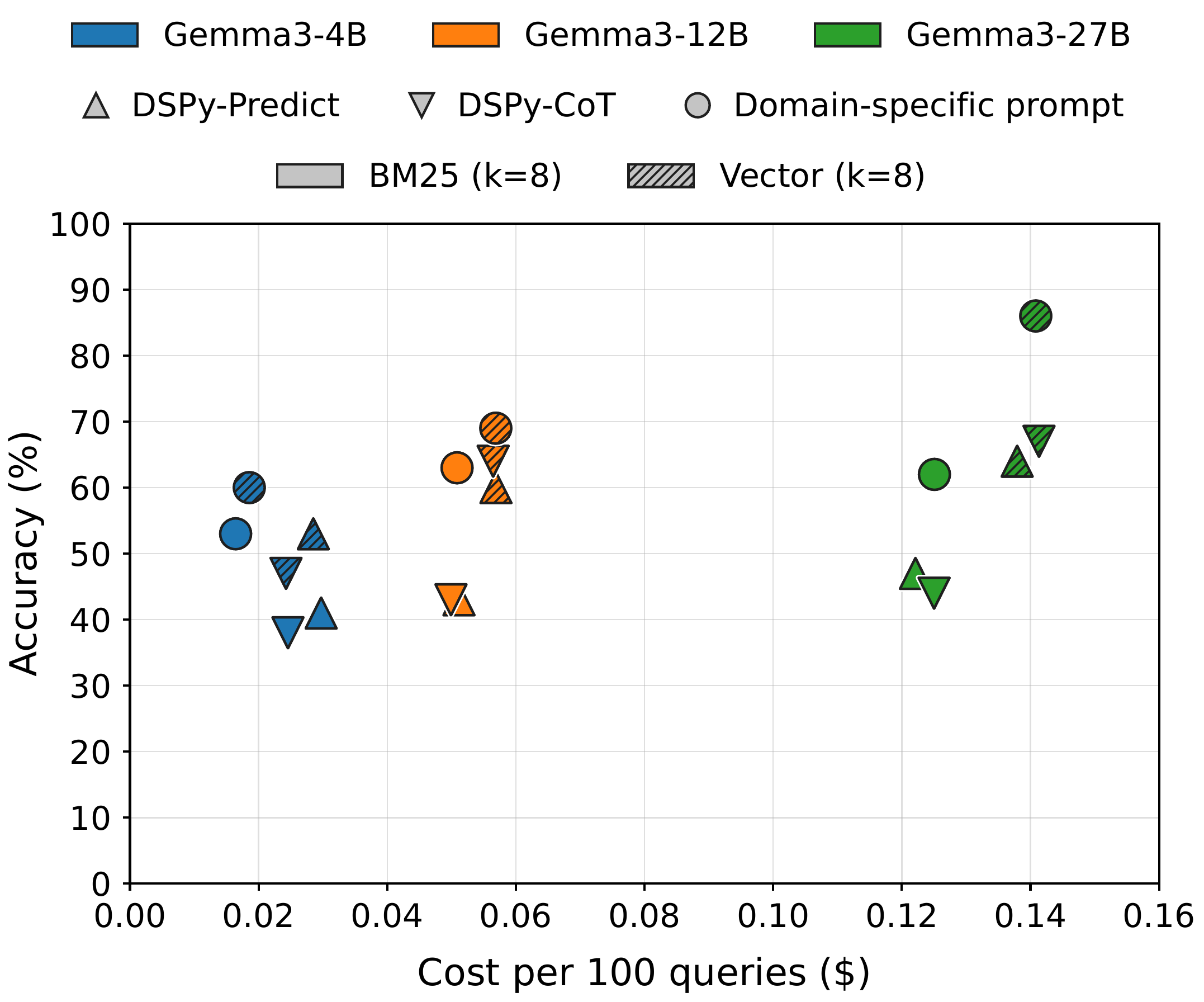}
\vspace{-3mm}
\caption{Accuracy vs. cost for each of the 18 configurations on 100 randomly chosen tasks from FinanceBench~\cite{financebench}.}
\vspace{-1mm}
\label{fig:credo-experiment}
\end{figure}

Imagine a financial analyst asking an AI assistant: ``What was the year-over-year revenue change for Apple in Q3 2023?''
To answer this question, the assistant must decide how much evidence to gather, which model to use for reasoning, and the best method for validating the result.
Most LLM orchestration frameworks (e.g., LangChain~\cite{langchain}) involve static control flow, either in the form of a fixed pipeline or explicit instructions embedded directly in the prompts.
This mixture of control flow with program logic makes the resulting behavior both rigid and difficult to reason about, especially when debugging.
Moreover, developers must carefully balance the ability to handle more complex tasks against the risk of overprovisioning for simpler ones.

To illustrate this problem, we randomly selected 100 tasks from FinanceBench~\cite{financebench} and ran them with 18 different pipeline configurations representing all combinations of three Gemma 3~\cite{gemma3} model sizes, three prompting strategies, and two retrieval methods.
The results, shown in \cref{fig:credo-experiment}, highlight two key findings.

First, retrieval method matters more than model size.
For example, with the 12B model and DSPy's~\cite{dspy} chain-of-thought (CoT) prompting strategy, simply switching the retrieval method from BM25 lexical search to vector similarity search increases accuracy from 43\% to 64\% at similar inference cost.
More strikingly, the 4B model with vector retrieval outperforms the 27B model with BM25 retrieval, even though the latter has a 5$\times$ higher inference cost.

Second, tuning the prompting strategy alone provides clear improvements for virtually all combinations of model size and retrieval method.
For instance, switching from DSPy-CoT to a domain-specific prompt with the 12B model and BM25 retrieval yields an accuracy improvement of 20 percentage points with no commensurate cost increase.
In some cases (e.g., with the 4B model and BM25 retrieval), this prompting strategy can even improve accuracy while simultaneously reducing cost.

This complex space of potential optimizations motivates the need for an adaptive approach that can react to runtime signals and select the appropriate execution flow on a case-by-case basis.
On one end of the spectrum are semantic query processing engines (e.g., LOTUS~\cite{lotus}, DocETL~\cite{docetl}, Palimpzest~\cite{palimpzest}), which expose composable LLM-powered operators for a more declarative approach to pipeline specification.
However, control flow is typically fixed when lowering these pipelines to corresponding physical plans, perhaps with minor adaptivity baked into the prompts of individual operators.
At the other extreme, agentic AI frameworks (e.g., ReAct~\cite{react}, AutoGen~\cite{autogen}) have no predefined control flow path and instead rely entirely on LLM reasoning to drive execution.
Agents offer substantial flexibility because they can reevaluate and modify current plans based on the results of each execution step, yet this autonomy makes precise behavioral control and enforcement of certain guarantees extremely challenging.

Our approach, called \sys, achieves the best of both worlds by allowing developers to define logical pipelines composed of high-level operators, with adaptive physical execution decisions handled separately by a database-backed semantic control plane.
\sys's core abstractions include \textit{beliefs}, which represent claims about runtime state (e.g., query complexity, evidence relevance), and \textit{policies}, which are declarative rules that prescribe how to adjust execution behavior in response to those beliefs.
As new runtime signals emerge, \sys's execution engine reevaluates beliefs, triggers policies, and updates the physical plan accordingly.
Consequently, these execution decisions become explicit, making runtime behavior more understandable and controllable.
In this demo, the audience will interactively compose beliefs and policies to adaptively control an LLM pipeline built for FinanceBench.
\section{Credo}
\label{sec:credo}


\begin{figure*}[ht!]
\centering
\includegraphics[width=0.93\textwidth,trim={3mm 70mm 4mm 4mm},clip]{figures/credo-system-1200dpi.png}
\vspace{-2mm}
\caption{Overview of \sysbf. Developers specify a logical LLM pipeline accompanied by a set of declarative beliefs and policies that form its semantic control plane, allowing adaptivity without directly modifying the underlying code. The observation layer evaluates beliefs based on available evidence, the decision layer evaluates policies over belief values and associated confidence scores, and the execution layer applies the corresponding rewrites before continuing with the updated physical plan.}
\vspace{-2mm}
\label{fig:credo-system}
\end{figure*}

We argue that the logical definition of LLM pipelines (i.e., the types and ordering of operators) should be decoupled from physical execution decisions (e.g., model choice, evidence retrieval method, result verification)~\cite{spear}.
As shown in \cref{fig:credo-system}, \sys achieves this separation through three complementary layers: (1) an \textit{observation layer} that makes runtime state explicit as high-level beliefs; (2) a \textit{decision layer} that manages declarative policies over those beliefs; and (3) an \textit{execution layer} that applies actions to adapt the pipeline when policies fire.
In the following, we describe each in more detail.

\begin{figure}[t]
\centering
\begin{lstlisting}[mathescape=true,linewidth=0.98\linewidth]
/* static */
BELIEF query_complexity(input:query):
    (value, confidence) = routellm_router(input:query)

/* dynamic */
BELIEF relevance(input:query, context:evidence):
    /* value $\in$ {correct, ambiguous, incorrect} */
    (value, confidence) = 
            llm_judge(input:query, context:evidence)
    DEPENDS ON {context:evidence}

/* operator rewrite */
WHEN query_complexity.value >= 0.25
     AND query_complexity.confidence >= 0.7
THEN REWRITE generate SET model = gemma3-27b

/* pipeline rewrite */
WHEN relevance.value = incorrect
     AND relevance.confidence >= 0.8
THEN INVALIDATE {evidence, model_answer, verified}
     REWRITE retrieve SET method = vector, top_k = 12
\end{lstlisting}
\vspace{-3mm}
\caption{Example belief and policy definitions.}
\vspace{-1mm}
\label{fig:beliefs-policies}
\end{figure}

\subsection{Observation Layer: Beliefs}
In the observation layer, \sys maintains a set of \textit{beliefs} that describe the current state, expressed as \texttt{(value, confidence)} pairs representing semantic claims and any associated uncertainty.
Beliefs are computed by extractors (e.g., heuristics, classifiers, LLM prompts) that either operate directly on raw evidence like documents or build on top of other beliefs.
\sys can resolve static beliefs before execution begins, whereas dynamic beliefs depend on intermediate results.

\cref{fig:beliefs-policies} shows two example beliefs.
The \texttt{query\_complexity} belief wraps an implementation of RouteLLM~\cite{routellm}, a lightweight LLM router that dynamically predicts whether a particular query will benefit from using a larger model.
Note that this belief is static, meaning that \sys can perform the extraction based solely on information available prior to operator execution.
The \texttt{relevance} belief, on the other hand, is dynamic, since it depends on information available only after retrieval (i.e., if retrieved documents are relevant for a particular query).
This belief is implemented as an LLM-as-a-judge extractor that evaluates whether the documents actually contain the information necessary to answer the question, returning a \texttt{(value, confidence)} pair where $\texttt{value} \in \{\texttt{correct}, \texttt{ambiguous}, \texttt{incorrect}\}$.

One obvious concern is the extraction cost, as some beliefs may be particularly expensive to evaluate (e.g., requiring LLM invocation).
To mitigate this problem, we are developing a cost-based optimizer for belief extraction planning, such that expensive beliefs are extracted as infrequently as possible.

\subsection{Decision Layer: Policies}
The decision layer handles how and when to adapt the pipeline based on the current set of beliefs.
Some existing frameworks support adaptivity in narrow circumstances, such as CRAG~\cite{crag} for retrieval quality or Self-Refine~\cite{self-refine} for answer quality.
In contrast, \sys externalizes this logic into \textit{policies}, which are declarative rules that map belief conditions to physical operator parameters.
Each policy specifies target operators and a set of these physical parameters to apply, with trigger conditions that can consider both the belief values and their associated confidence scores.
Developers can also assign an explicit priority to each policy, such that rewrites will occur in priority order if multiple policies fire at the same time.
Much like declaring an index to change physical access paths without rewriting a SQL query, \sys enables policy tuning independent of the pipeline's control flow; that is, developers can extend the adaptivity of LLM pipelines simply by updating the policy catalog rather than the underlying implementation.

\cref{fig:beliefs-policies} shows two example policies.
The first is an operator rewrite that fires when \texttt{query\_complexity.value~$\geq$~0.25} with sufficiently high confidence, modifying \texttt{generate} to use the largest available model.
The second is a pipeline rewrite that fires when \texttt{relevance.value = incorrect}, indicating that the retrieved evidence does not support an answer to the question.
The policy therefore invalidates existing state (i.e., \texttt{\{evidence, model\_answer, verified\}}) and rewrites \texttt{retrieve} to use vector retrieval with more candidates.
Note that these rewrites apply only to queries that trigger the policies, whereas those that do not will execute with the original pipeline.

\subsection{Execution Layer: Actions}
The execution layer applies the actions associated with policies whenever they fire and invokes their target operators.
Consistent with \sys's separation between logical pipeline definition and physical execution, operators have no direct knowledge of beliefs or policies; they simply receive a set of parameters and execute.
Instead, the semantic control plane handles all planning and optimization, similar to the traditional separation between logical relational operators (e.g., joins) and selection of a physical implementation (e.g., hash vs. sort-merge).

\sys currently provides three built-in operators: (1) \texttt{retrieve}, which gathers evidence using a configurable document retrieval strategy; (2) \texttt{generate}, which invokes the chosen LLM with a customizable prompting strategy; and (3) \texttt{verify}, which evaluates answer correctness via techniques like LLM-as-a-judge or chain-of-verification~\cite{cov}.
Developers can also register user-defined operators for their applications (e.g., custom API calls, domain-specific tools).
In the future, we plan to extend \sys's built-in operator set to include functionality like web search, code interpreters, and storage for long-term agentic memory.
\section{Demo Description}
\label{sec:demo}


\begin{figure*}[t]
\centering
\includegraphics[width=0.825\textwidth,trim={44mm 3mm 45mm 3mm},clip]{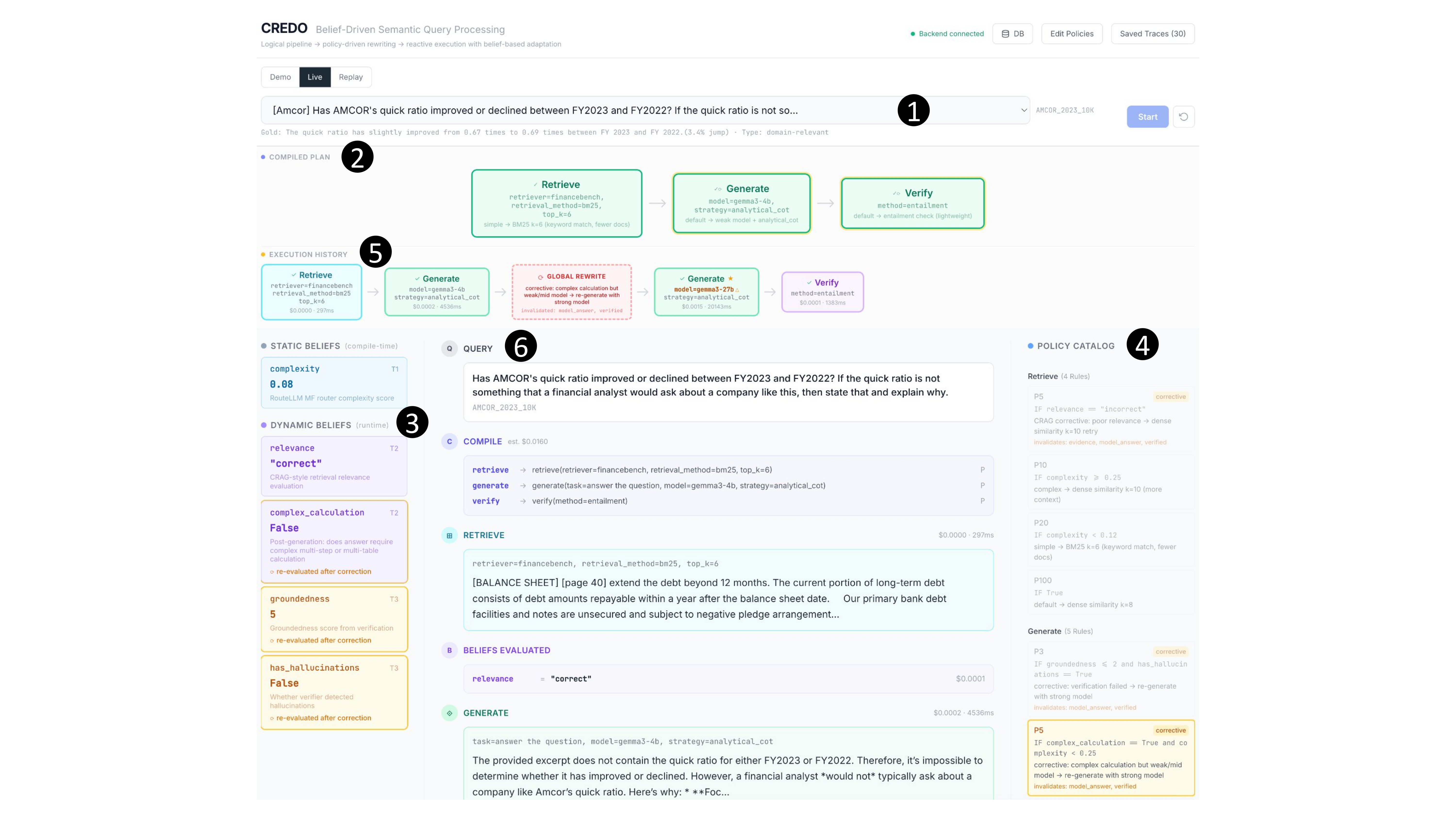}
\vspace{-2mm}
\caption{\sysbf's web interface.
\numcircledmod{1} \textit{Query Box:} Allows selecting from predefined benchmark tasks or submitting a custom query.
\numcircledmod{2} \textit{Logical Plan:} Shows the high-level operators that make up the LLM pipeline.
\numcircledmod{3} \textit{Belief Catalog:} Reports each belief's current \texttt{(value, confidence)} pair based on the available evidence and allows for the definition of new beliefs.
\numcircledmod{4} \textit{Policy Catalog:} Lists the active policies and supports inline updates, as well as the creation of new policies.
\numcircledmod{5} \textit{Physical Plan:} Displays the operator configurations produced during compilation and highlights subsequent rewrites as policies fire.
\numcircledmod{6} \textit{Event History:} Records a trace of evaluated beliefs, triggered policies, and operator executions, including details like inputs/outputs, cost, and latency.}
\vspace{-1mm}
\label{fig:credo-web}
\end{figure*}

The demo will center on FinanceBench~\cite{financebench}, a financial question answering benchmark with tasks ranging in difficulty from simple calculations to multi-step, cross-document reasoning.
The audience will use \sys's web interface (\cref{fig:credo-web}) to create custom beliefs and policies while observing their effects on pipeline execution.

As a jumping-off point, we will begin with two tasks from FinanceBench that appear superficially similar but demand very different execution strategies.
The first is a simple lookup that requires locating a single numerical value in an income statement: ``What was Apple's total revenue in FY2022?''
The second requires extracting two different numerical values from a table and then performing a two-step computation: ``What is Microsoft's FY2022 gross profit margin?''
As explained, a fixed pipeline would either waste resources on the former or risk errors on the latter.

Using these two questions, the audience will receive a brief tutorial that illustrates how \sys solves this problem through our framework of beliefs and policies.
First, we will register the RouteLLM~\cite{routellm} router from \cref{fig:beliefs-policies} as a belief, as well as specify corresponding retrieval and model choice policies (e.g., use the 4B model and BM25 retrieval with $k{=}6$ below a $0.12$ threshold).
We will also define a belief modeled on CRAG to evaluate whether the retrieved evidence is actually relevant to the question and related policy to switch to vector retrieval whenever this belief returns \texttt{incorrect}.
A similar \texttt{complex\_calculation} belief will determine if an answer requires combining multiple pieces of evidence, with a policy to regenerate using the 27B model and a modified prompt describing this requirement.
Finally, the \texttt{groundedness} and \texttt{has\_hallucinations} beliefs will assess whether the gathered evidence fails to support the conclusions and detect any potential hallucinations in the answer, respectively, either of which will trigger another corrective policy to retry the generation.

After the tutorial, the audience can freely explore \sys by submitting predefined tasks from the benchmark or their own custom queries via the query box \numcircledmod{1}.
The logical LLM pipeline consists of three steps (\texttt{retrieve} $\to$ \texttt{generate} $\to$ \texttt{verify}) and appears just below the query box \numcircledmod{2}.
The interface also displays the current belief catalog \numcircledmod{3} and policy catalog \numcircledmod{4}, which can be adjusted using the inline editor to see how changes affect the compiled plan.
Clicking \textit{Start} will compile and execute the pipeline, evaluating beliefs throughout execution and updating the physical plan as policies fire \numcircledmod{5}.
When the task completes, the interface will show the final answer along with details about every execution step \numcircledmod{6}, including a cost breakdown for each operator, trace of all belief evaluations and triggered policies, and summary of corrective rewrites applied.
\section{Conclusion \& Future Work}
\label{sec:future}

We presented \sys, a framework for controlling the runtime behavior of LLM pipelines through declarative beliefs and policies.
During the demonstration, the audience will observe firsthand how \sys can support runtime adaptivity for these pipelines without changing the underlying application code, all while making execution decisions more transparent.

Moving forward, we plan to extend \sys with new operators and tools to support a broader range of applications.
Our longer-term goal is to enable richer agentic workflows, potentially even automating the discovery of new beliefs and policies.
For example, a search process could propose candidates, test them against execution outcomes, and retain successful strategies for reuse across tasks, similar to approaches like Meta-Harness~\cite{metaharness}.
The resulting optimization loop would substantially reduce the burden on developers while keeping these decisions explicit.

\begin{acks}
This work was supported in part by a Brown Seed Grant and an award from Google's Research Scholar Program.
\end{acks}

\balance
\bibliographystyle{ACM-Reference-Format}
\bibliography{bib}

\end{document}